\pdfoutput=1

\documentclass[11pt]{article}

\usepackage[final]{acl}

\usepackage{times}
\usepackage{latexsym}

\usepackage[T1]{fontenc}

\usepackage[utf8]{inputenc}

\usepackage{microtype}

\usepackage{inconsolata}

\usepackage{graphicx}
\usepackage{multirow}
\usepackage{colortbl}
\usepackage{amsmath}
\usepackage{booktabs}
\usepackage{amssymb}
\usepackage{pifont}
\usepackage{enumitem}
\usepackage{todonotes}
\usepackage[most]{tcolorbox}
\usepackage{listings}

\usepackage{multicol}
\usepackage{lipsum}

\usepackage{xcolor}

\usepackage{tabularx}

\title{\includegraphics[height=1.5em]{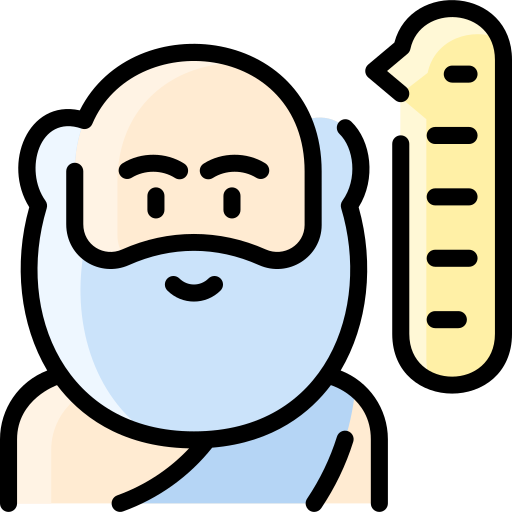} \textsc{Socratic-PRMBench}: Benchmarking Process Reward Models with Systematic Reasoning Patterns}

\author{
 \textbf{Xiang Li\textsuperscript{1,2,3}},
 \textbf{Haiyang Yu\textsuperscript{3}},
 \textbf{Xinghua Zhang\textsuperscript{3}},
 \textbf{Ziyang Huang\textsuperscript{1,2}},
 \\
 \textbf{Shizhu He\textsuperscript{1,2}\thanks{Corresponding author}} ,
 \textbf{Kang Liu\textsuperscript{1,2}},
 \textbf{Jun Zhao\textsuperscript{1,2}},
 \textbf{Fei Huang\textsuperscript{3}},
 \textbf{Yongbin Li\textsuperscript{3}}
\\
 \textsuperscript{1}Institute of Automation, Chinese Academy of Sciences
 \\
 \textsuperscript{2}School of Artificial Intelligence, University of Chinese Academy of Sciences
 \\
 \textsuperscript{3}Tongyi Lab, Alibaba Group
\\
}

\begin{document}
\maketitle
\begin{abstract}
Process Reward Models (PRMs) are crucial in complex reasoning and problem-solving tasks (e.g., LLM agents with long-horizon decision-making) by verifying the correctness of each intermediate reasoning step.
In real-world scenarios, LLMs may apply various reasoning patterns (e.g., decomposition) to solve a problem, potentially suffering from errors under various reasoning patterns. Therefore, PRMs are required to identify errors under various reasoning patterns during the reasoning process. 
However, existing benchmarks mainly focus on evaluating PRMs with stepwise correctness, ignoring a systematic evaluation of PRMs under various reasoning patterns. 
To mitigate this gap, we introduce \textsc{Socratic-PRMBench}, a new benchmark to evaluate PRMs systematically under six reasoning patterns, including \textit{Transformation},  \textit{Decomposition}, \textit{Regather}, \textit{Deduction}, \textit{Verification}, and \textit{Integration}. \textsc{Socratic-PRMBench} comprises 2995 reasoning paths with flaws within the aforementioned six reasoning patterns.
Through our experiments on both PRMs and LLMs prompted as critic models, we identify notable deficiencies in existing PRMs. 
These observations underscore the significant weakness of current PRMs in conducting evaluations on reasoning steps under various reasoning patterns. 
We hope \textsc{Socratic-PRMBench} can serve as a comprehensive testbed for systematic evaluation of PRMs under diverse reasoning patterns and pave the way for future development of PRMs\footnote{Our code and data are available at \url{https://github.com/Xiang-Li-oss/Socratic-PRMBench}}.

\end{abstract}

\section{Introduction}
\begin{figure}
    \centering
    \includegraphics[width=\linewidth]{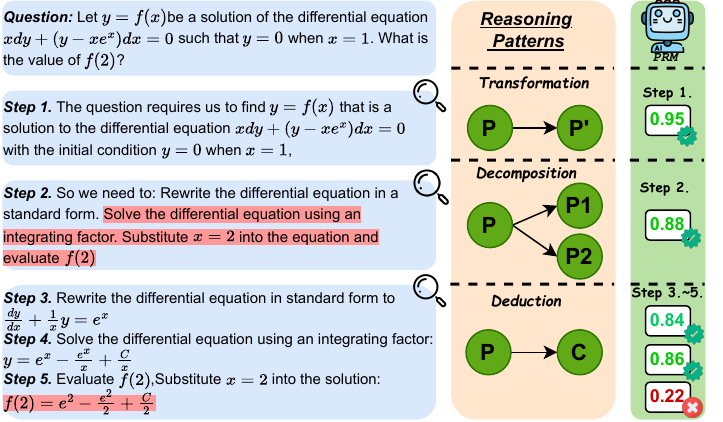}
    \caption{(Left): Given a question, the reasoning step 2 and 5 contain errors. (Medium): Each step applys a specific reasoning pattern. (Right): The process reward model successfully detects the error of \textit{Deduction} pattern but fails with the \textit{Decomposition} reasoning pattern. 
    }
    \label{fig:head_fig}
    \vspace{-10pt}
\end{figure}

\renewcommand{\arraystretch}{0.9} 
\newcommand{\tabincell}[2]{\begin{tabular}{@{}#1@{}}#2\end{tabular}}
\begin{table*}[ht]
  \centering
  \vspace{1mm}

  \resizebox{0.95\textwidth}{!}{
    \begin{tabular}{lccccccc}
    \toprule
     & \tabincell{c}{\textbf{PRM}\\ \textbf{Benchmarks?} } & \tabincell{c}{\textbf{Error Type}\\ \textbf{Detection?}}&  \tabincell{c}{\textbf{Fine-grained} \\ \textbf{Classes}} & \tabincell{c}{\textbf{Reasoning} \\ \textbf{Patterns$^\dagger$}}   & \textbf{Annotator} & \tabincell{c}{\textbf{Test Case}\\ \textbf{Size}} & \tabincell{c}{\textbf{Average}\\ \textbf{Steps}} \\
    \midrule
  
    RMBench \cite{liu2025rmbench}& \ding{55} & \ding{55} & 1 & 1 & Synthetic + Human & 1,327 & -\\
    CriticBench \cite{lin-etal-2024-criticbench}& \ding{55}& \ding{55} & 1 & 1  & - & - & -\\
    MathCheck-GSM \cite{zhou2025is}& \ding{55}  & \ding{55} & 1 & 1  & Synthetic & 516 & - \\
    ProcessBench \cite{zheng2024processbench}& \ding{51} & \ding{55} & 1 & 1 & Human & 3,400 & 7.1 \\
    PRMBench \citep{song2025prmbench}& \ding{51}& \ding{51}  & 9 & 1 & Synthetic + Human & 6,216 & 13.4 \\

    \midrule
    \textsc{Socratic-PRMBench}& \ding{51}& \ding{51}  & 20 & 6 & Synthetic + Human & 2995 & 8.7 \\ 
    \bottomrule
    \end{tabular}
  }
    \caption{Comparison between our proposed \textsc{Socratic-PRMBench} and other benchmarks or datasets for reward model evaluation.
  $^\dagger$:  the number of reasoning patterns covered within the benchmark. 
  }
  \vspace{-10pt }
  \label{tab:comparison}
\end{table*}

Large Language Models (LLMs) \citep{openai_o1_2024,deepseekr1,qwen-qwq-32b-preview} augmented by methodologies like Reinforcement Learning with Verifialble Rewards (RLVR) \citep{trung-etal-2024-reft,shao2024deepseekmath} and Test-Time Scaling \citep{snell2025scaling,bansal2025smaller}, have demonstrated significant capabilities in complex reasoning and decision-making tasks. Process Reward Models (PRMs) \citep{lightman2024lets, wang2023math,zhang2025lessons} play a crucial role in these advancements, especially for LLM agents which involve long-horizon decision-making steps~\cite{choudhury2025process,ma2025thinking,xiong2025rag}. By providing step-level rewards during the reasoning process, PRMs offer more accurate and denser reward signals, which in turn guide the optimization of LLMs and the exploration of reasoning trajectories \citep{tie2025survey,ji2025test}.

However, the diverse reasoning patterns applied by LLMs during reasoning process \citep{dong2023large,li2024mtmt} pose a challenge for PRMs in consistently providing accurate rewards.  
Figure \ref{fig:head_fig} illustrates such a scenario: according to the thoery of ancient Greek philosopher Socrate \citep{dong2023large,qi2023artsocraticquestioningrecursive}, the reasoning pattern for Step 1 is `\textit{Transformation}', for Step 2 `\textit{Decomposition}', and for Steps 3-5 `\textit{Deduction}'.
Although the existing PRM identifies the error in Step 5 (\textit{Deduction} pattern), it does not detect the fundamental cause of this error from the \textit{Decomposition} pattern. Specifically, in Step 2, the omission of substituting a point in the solution of the differential equation to calculate the constant $C$, causes $C$ to remain undetermined throughout the subsequent reasoning process, resulting in a flawed final answer. This observation indicates the unreliability of current PRMs towards diverse reasoning patterns.

For a comprehensive assessment of PRMs' error detection capabilities across various reasoning patterns, we introduce \textsc{\textsc{Socratic-PRMBench}}, a systematic and fine-grained benchmark. 
In contrast to prior benchmarks with limited systematic evaluation \citep{zheng2024processbench,song2025prmbench}, inspired by the ancient Greek philosopher Socrates, we design to evaluate PRMs' proficiency in detecting errors across 6 reasoning patterns: \textit{Transformation},  \textit{Decomposition}, \textit{Regather}, \textit{Deduction}, \textit{Verification}, and \textit{Integration}.
Specifically, \textsc{Socratic-PRMBench} comprises 2995 reasoning paths, with flaws categorized into six primary categories by reasoning pattern and 20 sub-categories of fine-grained error types.
The data annotation process for \textsc{Socratic-PRMBench} is fully automated using LLMs, thereby obviating the need for extensive human labor. 
We ensure the difficulty of the data through rule-based filtering and guarantee its quality through manual expert review.

We conducted extensive experiments on a wide range of models, including open-source PRMs, and a series of general-purpose and reasoning-specialized LLMs. 
The findings reveal considerable scope for improvement in current PRMs. Notably, Qwen2.5-Math-PRM, the highest-performing PRM, attained a mere 68.0 overall score.
Through detailed analytical experiments, we identified substantial disparities in the error detection capabilities of current PRMs across different reasoning patterns, alongside evident latency in indentifying error steps and significant bias of reward generation.
By leveraging \textsc{Socratic-PRMBench} for evaluation, we offer a pathway to comprehensively assess PRMs from the perspective of reasoning patterns. This can be potentially helpful for mitigating the risk of reward hacking in future PRM development. 
In general, our contributions are summarized as follows:

\begin{itemize}
\item We propose \textsc{Socratic-PRMBench}, the first systematic PRM benchmark from reasoning pattern perspective, comprising 2995 samples for a comprehensive and fine-grained evaluation on process reward models.

\item Based on ancient Greek logic theory \citep{qi2023artsocraticquestioningrecursive}, \textsc{Socratic-PRMBench} covers 6 carefully designed reasoning patterns, including \textit{Transformation},  \textit{Decomposition}, \textit{Regather}, \textit{Deduction}, \textit{Verification}, and \textit{Integration}, with 20 sub-categories of fine-grained error types.
This systematic and granular evaluation framework enables a comprehensive assessment of PRMs and facilitates the identification of their potential shortcomings.

\item We perform extensive experiments on a wide range of SOTA PRMs and LLMs with \textsc{Socratic-PRMBench}. Our results reveal essential limitations in current PRMs and offer insights for future progress in this area.

\end{itemize}

\section{Related Work}

\paragraph{Process Reward Models}
Process reward models (PRMs) have demonstrated their superiority over outcome reward models (ORMs) \citep{zhang2024generative,ankner2024critiqueoutloudrewardmodels} by providing more accurate and dense reward signals for intermediate reasoning steps.
As a result, the development of PRMs is gaining increasing attention.
\citet{lightman2024lets} contributes a manually annoted dataset for PRM training, \citet{wang-etal-2024-math} propose an automatic step-level labeling method with Monte Carlo estimation.
Moreover, \citet{dong2024rlhf, zhao2025genprm} forms process reward modeling as generation task and improve generative capabilities of PRMs using CoT reasoning.
In contrast to the flourish of PRMs' training, PRMs' evaluation remaines comparatively underdeveloped. To remedy this imbalance, we present \textsc{Scoratic-PRMBench}, a novel benchmark for PRMs' evaluation.

\paragraph{Reward Model Benchmarks}
Reward benchmarks are crucial for evaluating reward models, as they provide a direct and quantifiable measure. 
Despite the emergence of numerous benchmarks \citep{liu2025rmbench,lin-etal-2024-criticbench,lambert2024rewardbench}, they are are primarily designed to evaluate ORMs, without 
any step-level annotations.
\citet{zheng2024processbench, song2025prmbench}  annotate step-level labels using LLMs and human experts to create benchmarks for PRMs. However, their evaluation are not systematic and ignore the need to evaluate PRMs' error detection capabilities towards diverse reasoning patterns \citep{dong2023large,li2024mtmt}.
To address this gap, we propose \textsc{Socratic-PRMBench}, a systematic and granular benchmark to provide a comprehensive assessment of PRMs from the perspective of reasoning patterns. A comparison between our \textsc{Socratic-PRMBench} and existing reward model benchmarks is summarized in Table \ref{tab:comparison}.

\section{Socratic-PRMBench}
\label{sec:method}
\subsection{Reasoning Patterns}
\label{sec:pattern}
\begin{figure*}
    \centering
    \includegraphics[width=0.95\linewidth]{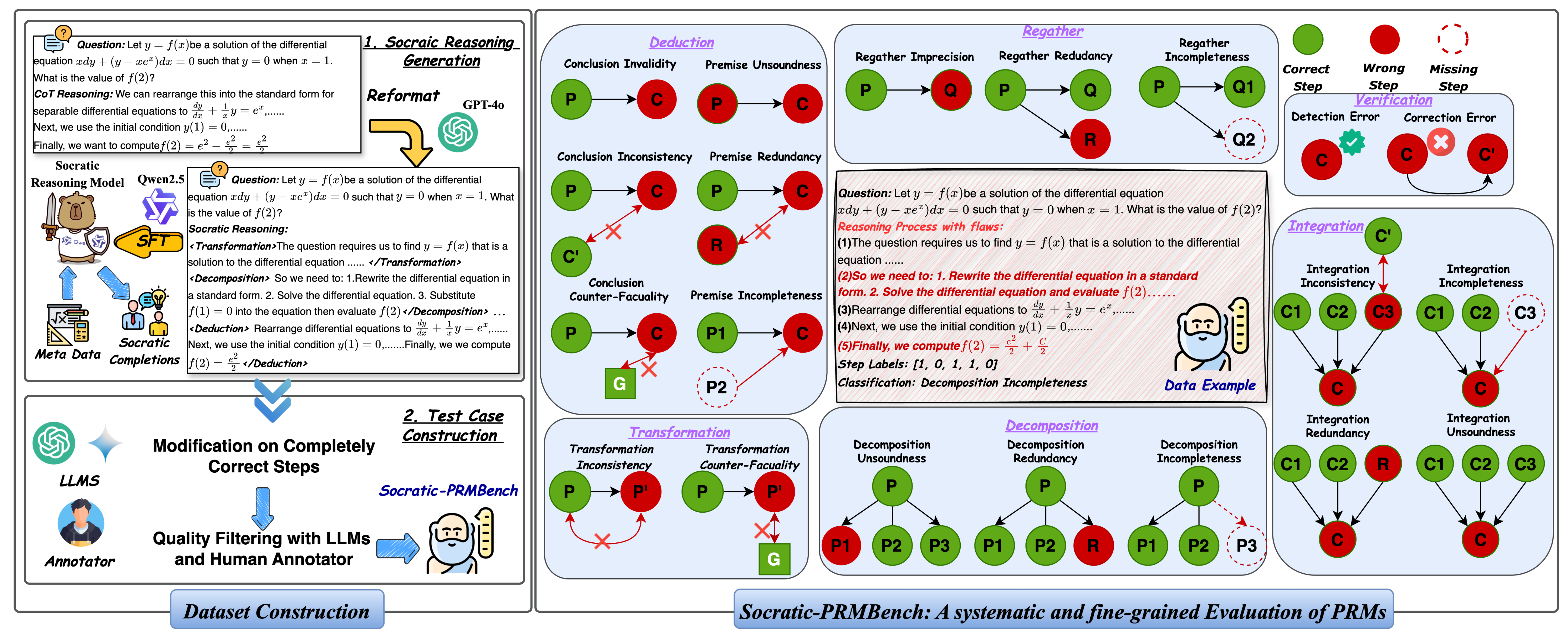}
    \caption{An overview of our \textsc{Socratic-PRMBench}. The left part illustrates our dataset constuction procedure. The right part illustrates the 6 reasoning patterns and 20 sub-categories of fine-grained error types. We use $P$ and $C$ to represent (sub)problems and conclusions, respectively. We use $Q$, $R$, $G$ to represent gathered information, redundant contents, and ground truth. 
    } 
    \vspace{-10pt}
    \label{fig:main_fig}
\end{figure*}

The design of the reasoning patterns in \textsc{Socratic-PRMBenchmark} is inspired by the logical theories of the ancient Greek philosopher Socrates. As Socrates once stated, "I cannot teach anybody anything. I can only make them think." Following this philosophical wisdom, we categorize reasoning into six atomic reasoning patterns, within these six reasoning patterns, we systematically design a total of 20 types of reasoning errors. The atomic reasoning patterns and the fine-grained categories of error types under Socrates' logical framework are illustrated in Figure \ref{fig:main_fig}.

\noindent \textbf{Transformation} transforms the problem into a homogeneous or similar problem, or abstract the problem. It usually explains the problem from a problem-solving perspective, aiming at gaining a more comprehensive and clear understanding of the problem. Specifically, the \textit{Transformation} evaluation category can be divided into two sub-categories: \textbf{\textit{Transformation Inconsistency}} and \textbf{\textit{Transformation Counter-Factuality}}. For a \textit{Transformation} step $P \rightarrow P'$, Transformation Inconsistency refers that $P'$ lacks consistency in logic, semantics, or understanding with $P$.
Transformation Counter-Factuality refers to including factual error that against ground truth $G$ in $P'$.

\noindent \textbf{Decomposition} breaks the problem into manageable subproblems, or makes a plan for reasoning steps, resolving the main problem by tackling each sub-problem. 
Specifically, the \textit{Decomposition} evaluation category can be divided into three sub-categories: \textbf{\textit{Decomposition Unsoundness}}, \textbf{\textit{Decomposition Redundancy}}, and \textbf{\textit{Decomposition Incompleteness}}. For a \textit{Decomposition} step $P\rightarrow\{P_1, P_2,...,P_n\}$, each of the three sub-categories represents a distinct type of error in subproblem $P_i$, which can be incorrect caused by logical inequality, missing important sub-problems and conditions, or including redundant sub-problems and constrains.

\noindent \textbf{Regather} collects key information from the input relevant to solving the problem and identifies crucial principles, and other concepts related to solving the problem.
Specifically, the \textit{Regather} evaluation category can be divided into three sub-categories: \textbf{\textit{Regather Imprecision}}, \textbf{\textit{Regather Redundancy}}, and \textbf{\textit{Regather Incompleteness}}.
For a \textit{Regather} step $P\rightarrow\{Q_1,Q_2,...,Q_n\}$, 
Regather imprecision refers to collecting a $Q_i$ with misinformation, misusing definations that are not suitable for solving the problem $P$. 
Regather Redundancy gathers redundant or unrelevant information not related with $P$.
Regather Incompleteness refers to the absence of core definations, critical principles and concepts.

\noindent \textbf{Deduction} derives a conclusion for a given premise directly. 
Specifically, the \textit{Deduction} evaluation category can be divided into six sub-categories: \textbf{\textit{Premise Unsoundness}}, \textbf{\textit{Premise Incompleteness}}, \textbf{\textit{Premise Redundancy}}, \textbf{\textit{Conclusion Invalidity}}, \textbf{\textit{Conclusion Inconsistency}} and \textbf{\textit{Conclusion Counter-Factuality}}.
For a \textit{Deduction} step $P\rightarrow C$, the first three sub-categories arise from the premise and include: (1) starting deduction resoning from an unreasonable or incorrect premise, (2) introducing redundant assumptions into the premise, and (3) omitting key conditions and constraints. 
The remaining three sub-categories originate from the conclusion and include: (1) deriving an invalid conclusion from correct premises, (2) deriving a conclusion that contradicts a previous conclusion, and (3) deriving a conclusion that is inconsistent with known ground truth.

\noindent \textbf{Verification} examines reasoning steps in terms of factual accuracy, logical consistency, etc, detecting potential errors and refining them iteratively. 
Specifically, the \textit{Verification} evaluation category can be divided into two sub-categories: \textbf{\textit{Detection Error}} and \textbf{\textit{Correction Error}}.
The former refers to failing to identify an incorrect conclusion $C$, The latter, however, involves recognizing the initial error in $C$ but introducing a new error during the attempted correction, leading to a different, incorrect conclusion $C'$.

\noindent \textbf{Integration} summarizes concluded conclusions to derive a new conclusion, integrating all current reasoning processes to form the final conclusion.
Specifically, the \textit{Integration} evaluation category can be divided into four sub-categories: \textbf{\textit{Integration Inconsistency}}, \textbf{\textit{Integration Incompleteness}}, \textbf{\textit{Integration Redundancy}}, and \textbf{\textit{Integration Unsoundness}}.
For an integration step $\{C_1,C_2,...,C_n\}\rightarrow C$,
the first three error types originate from a intermediate conclusion $C_i$, including the presence of conclusions that contradicts prior findings,the absence of crucial conclusions, and the introduction of unnecessary or redundant conclusions. 
The final error type, namely Integration Unsoundness, refers to concluding a final conlusion $C$ that is incorrect or unreasonable, even when integrated conclusions all satisfy soundness an completeness.

\begin{table*}[t]
  \centering
  \fontsize{13}{13.5}\selectfont

  \resizebox{1.0\textwidth}{!}{
    \begin{tabular}{l c cccccc}
    \toprule
    	&\textbf{Overall}& \textbf{\textit{Transformation}}&\textbf{\textit{Decomposition}}&\textbf{\textit{Regather}}&\textbf{\textit{Deduction}}&\textbf{\textit{Integration}}&\textbf{\textit{Verification}}\\
        \midrule
            
            Avg. Steps&8.7&8.5&8.7&8.6&8.5&8.5&10.8\\
            Avg. Error Steps&3.0&4.2&3.3&2.9&3.0&2.0&3.8\\
            Avg. First Error Step&4.7&1.5&3.0&3.1&5.4&7.2&6.9\\
            Avg. Question Length&209.6&224.4&220.7&207.5&221.7&191.3&169.4\\
            \# of Instances&2995&313&463&463&926&615&215\\

        \bottomrule
    \end{tabular}
  }
    \caption{Statistics of \textsc{Socratic-PRMBench}. 
  }
  \label{tab:data_statistics}
\end{table*}

\subsection{Benchmark Construction}
The dataset construction pipeline comprises two core stages: \textbf{Socratic Reasoning Generation} and \textbf{Test Case Construction}.

\subsubsection{Socratic Reasoning Generation}
This stage aims to create a data pool of Socratic reasoning process, represented as a sequence of atomic Socratic reasoning actions. As illustrated in left part of Figure \ref{fig:main_fig}, 
each reasoning step is enclosed with a start tag <[Pattern]> and an end tag </[Pattern]>. 
The content within the [Pattern] placeholder indicates the specific reasoning pattern that characterizes this particular step.

\paragraph{Socratic Reasoning Model Training} Given the scarcity of available Socratic reasoning data, we initially trained a specialized Socratic reasoning model to facilitate data generation. 
To achieve this, we sample 19k instances from the MATH-Hard \citep{hendrycks2021measuring} and Open-o1 \citep{openo1} datasets and prompt GPT-4o to transform their existing Chain-of-Thought (CoT) annotations into Socratic reasoning processes.
We then fine-tune Qwen2.5-72b-instruct \citep{qwen2.5} on these Socratic reasoning processes, yielding our Socratic reasoning model, denoted as $M_{Socratic}$.

\paragraph{Socratic Reasoning Generation}Subsequently, we leverage $M_{Socratic}$ to generate new Socratic reasoning processes from metadata. 
To this end, we first collect samples from GSM8k \citep{cobbe2021training}, Omni-Math \citep{omnimath}, MathBench \cite{liu-etal-2024-mathbench}, and OlympiadBench \citep{he-etal-2024-olympiadbench}.
In order to ensure that our problems are adequately challenging, we carefully curated the Omni-Math and MathBench datasets. 
Specifically, we excluded any Omni-Math samples with a difficulty rating lower than 4.0. 
For MathBench, we focused solely on MathBench-A, as this subset emphasizes theoretical application rather than conceptual understanding. 
Furthermore, we only retained instances from MathBench-A that are designated as high school or university level.
This procedure finally results in a data pool $D$.
For each question-answer pair $(q_i, a_i)$ in $D$, $M_{Socratic}$ generates a Socratic reasoning process $r_i$, resulting in a $(q_i, r_i, 
\hat{a_i})$ triplet.

\paragraph{Socratic Reasoning Curation}Finally, each $(q_i, r_i, \hat{a_i})$ tuple undergos a rigorous dual verification process: answer correctness was first assessed, followed by LLM-based verification of each individual step. 
Only tuples that pass both verifications are retained, resulting in our curated metadata set $D'$.
For answer verification, we follow Qwen2.5-Math \citep{qwen25math}, requiring that the predicted answer $\hat{a_i}$ satisfies both numerical and symbolic equivalence with the ground truth answer $a$. 
For step verification, we leverage GPT-4o \citep{openai2024gpt4o} to assess the correctness of each individual step in the reasoning process, with the detailed prompt in Appendix \ref{appendix_prompt}.

\subsubsection{Test Case Constuction}
In this stage, we generate test sets for each error type $C$ (as classified in Section \ref{sec:pattern}) by employing a controlled error injection procedure. 
For each error type $C$ (e.g., Repeat Inconsistency), we create a test set $T_C$.
This is achieved by first randomly select $N$ samples from the metadata set $D'$. 
And then for each sample $(q_i,r_i,a_i)$, including a problem $q_i$, a Socraitc reasoning path $r_i$  guaranteed completely correct through our dual verification process, we prompt GPT-4o to modify the originally correct reasoning process $r_i$, intentionally introducing an error consistent with error type $C$:
\begin{equation} 
\begin{aligned}
&\tilde{r}_i = \mathrm{LLM}(I,[q_i,r_i,a_i],C) \\
&T_C = \{t_i = (q_i,\tilde{r}_i,\tilde{a}_i)\}_{i=1}^N
\end{aligned}
\end{equation} 
where $\tilde{r}_i$ is the modified socratic reasoning process with the type of error $C$ and $I$ is the instruction prompt for GPT-4o to modify original process $r_i$ to $\tilde{r}_i$, with detailed prompt in Appendix \ref{appendix_prompt}.

\begin{table*}[t]
\belowrulesep=0pt
\aboverulesep=0pt
\fontsize{11}{13.5}\selectfont

\centering

\resizebox{0.96\textwidth}{!}{

\begin{tabular}{l cc| ccc| ccc | cc}
\toprule[1.5pt]
\multirow{2}{*}{\textbf{Model}}  & \multicolumn{2}{c|}{\textbf{Transformation}}  & \multicolumn{3}{c|}{\textbf{Decomposition}}& \multicolumn{3}{c|}{\textbf{Regather}} & \multicolumn{2}{c}{\textbf{Verification}}\\
\cmidrule(lr){2-3} \cmidrule(lr){4-6} \cmidrule(lr){7-9} \cmidrule(lr){10-11} 
& \textbf{TT.} & \textbf{TF.} & \textbf{DC.} &\textbf{DR} &\textbf{DS.}&\textbf{GP.} &\textbf{GC.} & \textbf{GR.} &\textbf{CE.} & \textbf{DE.}   \\
 \midrule
\hline &\multicolumn{10}{c}{\textit{\textbf{Process Reward Models (PRMs)}}} \\   \hline 
\href{https://huggingface.co/Skywork/Skywork-o1-Open-PRM-Qwen-2.5-7B}{Skywork-PRM-7B}  & 38.7	& 38.4 & 42.7	& 42.5 & 38.0 & 42.8 & 44.8	& 41.3 & 47.9 & 46.7 \\
\href{https://huggingface.co/GAIR/ReasonEval-7B}{ReasonEval-7B}  & 50.9	& 50.9 & 59.3 & 50.1 & 53.7 & 52.4&59.6&49.7&66.7&59.2 \\

\href{https://huggingface.co/RLHFlow/Llama3.1-8B-PRM-Mistral-Data}{RLHFlow-PRM-Mistral-8B}  & 50.6 &	52.7 & 46.6	& 47.3 & 42.7  & 38.0 & 44.6 & 48.7 & 53.1 & 49.5 \\
\href{https://huggingface.co/RLHFlow/Llama3.1-8B-PRM-Deepseek-Data}{RLHFlow-PRM-Deepseek-8B} & 47.5&	50.8&50.6&50.9&44.0&41.6&48.6&55.4&45.9&47.6 \\

\href{https://huggingface.co/peiyi9979/math-shepherd-mistral-7b-prm}{MathShepherd-Mistral-7B} & 54.5 & 50.9 & 59.4	& \textbf{57.4}	& 56.7 & \textbf{60.9} & 59.4 & 54.6 & \textbf{72.7} & \textbf{72.1} \\
\href{https://huggingface.co/Qwen/Qwen2.5-Math-PRM-7B}{Qwen2.5-Math-PRM-7B} & \textbf{55.8} & \textbf{64.3} & \textbf{61.7} & 51.6 & \textbf{58.4} & 57.5 & \textbf{61.8} & \textbf{58.2} & 67.4 & 64.1 \\

\hline &\multicolumn{10}{c}{\textit{\textbf{LLMs, Prompted as Critic Models}}} \\   \hline 
\href{https://openai.com/index/hello-gpt-4o/}{GPT-4o}&\textbf{62.4}&60.5&69.9&60.0&66.1&64.9&\textbf{74.1}&57.9&74.4&75.8 \\

\href{https://github.com/deepseek-ai/DeepSeek-R1}{Deepseek-R1}&51.9&\textbf{72.6}&63.4&64.4&67.1&70.9&	64.6&54.8&75.0&77.1 \\

\href{https://huggingface.co/Qwen/QwQ-32B-Preview}{QwQ-32B}&60.2&68.6&70.0&\textbf{67.9}&59.8&73.7&65.8	&55.4&75.8&75.7\\

\href{https://deepmind.google/technologies/gemini/pro/}{Gemini-2.5-Pro}&62.3&64.4&67.3&61.4&\textbf{68.5}&70.2&69.2&\textbf{58.6}&\textbf{78.3}&\textbf{78.0}\\

\href{https://openai.com/index/openai-o3-mini/}{o3-mini}&\textbf{62.4}&67.4&\textbf{70.4}&57.3&68.0&\textbf{77.3}&71.3&53.0&77.2&72.6\\

\bottomrule[0.8pt]

\end{tabular}}

\vspace{1em}

\resizebox{0.96\textwidth}{!}{
\begin{tabular}{l c|cccccc|cccc}
\toprule[0.8pt]
\multirow{2}{*}{\textbf{Model}} & \multirow{2}{*}{\textbf{Overall}} & \multicolumn{6}{c|}{\textbf{Deduction}}  & \multicolumn{4}{c}{\textbf{Integration}}\\
\cmidrule(lr){3-8} \cmidrule(lr){9-12} 
&& \textbf{CF.} & \textbf{CT.} & \textbf{CV.} &\textbf{PC.} &\textbf{PR.}&\textbf{PS.} &\textbf{IC.} & \textbf{IT.} &\textbf{IR.} & \textbf{IS.}   \\
 \midrule

\hline &\multicolumn{11}{c}{\textit{\textbf{Process Reward Models (PRMs)}}} \\   \hline 
\href{https://huggingface.co/Skywork/Skywork-o1-Open-PRM-Qwen-2.5-7B)}{Skywork-PRM-7B} & 43.6& 42.5&41.2&40.0&41.8&42.8&39.8&38.7&42.6&39.4&44.2 \\

\href{https://huggingface.co/GAIR/ReasonEval-7B}{ReasonEval-7B} & 61.9& 63.6&	63.6&66.3&61.9&65.2&63.5&69.7&78.2&68.7&76.1 \\

\href{https://huggingface.co/RLHFlow/Llama3.1-8B-PRM-Mistral-Data}{RLHFlow-PRM-Mistral-8B} & 48.8 & 50.4&46.2&45.2&46.1&44.5&43.3&51.2&58.1&46.6&56.3 \\
\href{https://huggingface.co/RLHFlow/Llama3.1-8B-PRM-Deepseek-Data}{RLHFlow-PRM-Deepseek-8B} & 51.5&51.5&52.4&52.0&47.6&51.4&45.2&55.3&63.7&53.3&66.7 \\

\href{https://huggingface.co/peiyi9979/math-shepherd-mistral-7b-prm}{MathShepherd-Mistral-7B} & 64.4 & 68.0&65.9	&66.5&62.4&65.9&65.4&63.1&74.2&60.1&72.3 \\
\href{https://huggingface.co/Qwen/Qwen2.5-Math-PRM-7B}{Qwen2.5-Math-PRM-7B} & \textbf{68.0} & \textbf{74.7}&\textbf{73.1}&\textbf{72.2}&\textbf{66.6}&\textbf{72.4}&\textbf{67.2}&\textbf{75.0}&\textbf{85.2}&\textbf{69.6}&\textbf{86.9} \\

\hline &\multicolumn{11}{c}{\textit{\textbf{LLMs, Prompted as Critic Models}}} \\   \hline 
\href{https://openai.com/index/hello-gpt-4o/}{GPT-4o} & 70.8&63.6&62.7&74.5&73.2&60.1&76.1&73.4&	80.8&52.7&88.7\\
\href{https://github.com/deepseek-ai/DeepSeek-R1}{Deepseek-R1}&73.0&80.8&72.6&77.2&68.6&72.0&76.9&75.9&78.9&59.9&88.6 \\

\href{https://huggingface.co/Qwen/QwQ-32B-Preview}{QwQ-32B}&73.8&70.3&75.0&\textbf{85.2}&\textbf{74.0}&69.5&77.5&\textbf{81.8}&83.5&58.7&96.7\\

\href{https://deepmind.google/technologies/gemini/pro/}{Gemini-2.5-Pro}&73.5&72.8&77.7&83.5&69.0&65.9&73.5&73.2&\textbf{88.9}&56.9&\textbf{96.9}\\

\href{https://openai.com/index/openai-o3-mini/}{o3-mini}&\textbf{75.7}&\textbf{83.3}&\textbf{81.0}&81.4&73.9&\textbf{75.3}&\textbf{78.6}&78.7&87.3&\textbf{72.0}&87.0\\

\bottomrule[1.5pt]

\end{tabular}}

\caption{
Evaluation results on \textsc{Socratic-PRMBench}. (Up): The PRM-Score of \textit{Transformation}, \textit{Decomposition}, \textit{Regather}, and \textit{Verification}. (Down): The PRM-Score of \textit{Deduction}, \textit{Integration} and \textit{Overall} performance. The best performance for each category and task is in \textbf{bold}. The full names of abbreviations are shown in Appendix \ref{sec:appendix}}
\vspace{-10pt}
\label{tab:main_results}
\end{table*}

\subsection{Quality Control}

To ensure the high quality and reliability of \textsc{Socratic-PRMBench}, we utilize both rule-based and LLM-based method 
to filter out any unsuitable samples, thereby ultimately creating our \textsc{Socratic-PRMBench}.

\paragraph{Rule-based Fitering}
Despite providing detailed task descriptions and output format requirements in the instruction $I$, GPT-4o may still occasional fail to follow the instruction $I$ strictly. 
Therefore, we implement a rule-based filtering method.
First, we use string matching to identify and remove any sample that fails to produce output in JSON format, which is required in $I$.
Second, we use regular expression to discard any sample that fail to successfully output the final answer.

\paragraph{LLM-based Filtering}
To ensure the quality of our generated test cases, we employ Gemini2.5-Pro to evaluate each sample $(q_i,\tilde{r}_i,\tilde{a}_i)$ , within a test set $T_C$ for a given error type $C$. 
Specifically, we instruct Gemini2.5-Pro to assess the sample based on two criteria: (1) the reasoning path $\tilde{r}_i$ appears superficially plausible yet contains an underlying reasoning error, and (2) the identified error should definitively belong to the targeted error type $C$, with detailed prompt shown in Appendix \ref{appendix_prompt}.  
After filtering by Gemini2.5-Pro, the acceptance rate of samples reaches 92.7\%, and 2995 samples are retained to form the final Socratic-PRMBench. The statistics of Socratic-PRMBench are shown in Table \ref{tab:data_statistics}.

\paragraph{LLM's Consistency with Human Annotators}
To demonstrate Gemini2.5-Pro's ability to perform this quality filtering task, we measure its agreement with human annotators.
We recruit three volunteer annotators, each holding at least a bachelor's degree, and ask them to verify a randomly sampled 10\% subset of our data using the exact same criteria with Gemini2.5-Pro. 
We then calculate the agreement rate between Gemini2.5-Pro and the human annotators. 
As a result, Gemini2.5-Pro shows a high degree of consistency with the human annotators, achieving an average agreement rate of 93.3\%.
This high level of consistency provides strong evidence that Gemini2.5-Pro can effectively replace human annotators in performing quality filtering across the entire dataset, reducing the burden of extensive manual work.

\section{Experiments}

\subsection{Models}
In our setting, we consider two types of model: Process Reward Models (PRMs) and Large Language Models (LLMs) prompted as critic models.

\paragraph{Process Reward Models (PRMs)} are trained with annotations of intermediate reasoning steps to evaluate and supervise intermediate reasoning process of language models. 
Our evaluation includes state-of-the-art open-source PRMs, such as:
(1) MathShepherd \citep{wang2023math}, which obtains the process label for each step by estimating the empirical probability of that step leading to the correct final answer.
(2) Two LLaMA-3.1-based Generative PRMs \citep{dong2024rlhf} that determine correctness based on the output probabilities of "Yes/No" tokens.
(3) ReasonEval \citep{mondorf2024beyond}, which asseses redundancy in addition to validity of reasoning steps.
(4) Two PRMs trained on the popular mathematical model Qwen2.5-Math, namely Skywork-PRM \citep{skyworkopeno12024} and Qwen2.5-Math-PRM \citep{zhang2025lessons}.

\paragraph{Large Language Models (LLMs) Prompted as Critic Models} Critic models aim to provide feedback and critique directly on model-generated texts, harnessing the generative power of Large Language Models. Our evaluation includes both general-purpose models, including GPT-4o \citep{openai2024gpt4o}, Gemini2.5-Pro \citep{gemini2.5pro}, and models specilized on reasoning, including Deepseek-R1 \citep{deepseekr1}, QwQ-32B \citep{qwen-qwq-32b-preview}, and o3-mini \citep{o3mini}.

\subsection{Evaluation Metrics}
Given that the evaluation of PRM centers on the detection of flawed reasoning steps, a straightforward application of Accuracy or F1-score may be affected by inherent biases of models. 
To address this concern, we follow \citep{song2025prmbench,zheng2024processbench} and employ the PRM-score as our evlauation metric, defined formally as:
\begin{equation} \label{eq1}
\begin{aligned}
\mathtt{\rm PRM}\mbox{-}\mathtt{\rm Score} &= w_1 \times {\rm F1}_{\rm neg} + w_2 \times {\rm F1}
\end{aligned}
\end{equation}
where F1 and F1$_{\rm neg}$ refer to F1 scores and negative F1 scores. $w_1$ and $w_2$ are weights that balance the contributions of the F1-score and negative F1-score. 
Following previous studies \citep{song2025prmbench,zheng2024processbench}, we set $w_1=w_2=0.5$.

\subsection{Main Results}
Our evluation results are exhibited in Table \ref{tab:main_results}. Our findings are as follow:

\paragraph{Comparision between PRMs and LLMs}The performance of PRMs is demonstrably inferior to that of LLMs. The top-performing PRM, Qwen2.5-Math-PRM-7B, achieves a score of only 68.0, which is lower than even the least effective LLM, GPT-4o. Furthermore, some PRMs perform below the level of random guess, highlighting their limitations in handling reasoning errors across diverse reasoning patterns. This suggests a considerable gap between PRMs and LLMs, indicating a need for substantial improvement. The challenges of PRM data annotation and the difficulty in ensuring the quality of synthetic data likely contribute to this disparity. For instance, Math-shepherd leverages synthetic data where step correctness is measured based on the estimated probability of arriving at the correct final answer, whereas Qwen2.5-Math-PRM-7B uses the manually labeled PRM800k dataset.

\paragraph{Comparision among LLMs}In contrast to PRMs, LLMs exhibit the potential to provide more robust and reliable rewards in critique, owing to their sophisticated language and reasoning skills. Consistent with this, we observe that reasoning-specialized LLMs outperforms general-purpose LLMs. Notably, QWQ-32B performs best among the open-source models and even outperforms GPT-4o. While QWQ -32B demonstrates impressive performance, it still underperforms o3-mini, indicating that although the gap in problem-solving performance is getting closer between open-source and proprietary models, a significant gap persists in their capabilities as critic models.

\paragraph{Redundant Errors Are More Challenging}We observed notable performance variations across fine-grained error types, even within the same reasoning pattern. Redundant errors, such as decomposition redundancy, regather redundancy, and integration redundancy within the Decomposition, Regather, and Integration patterns, consistently posed a greater challenge for both PRMs and LLMs compared to other error types within the same reasoning pattern. 
This may be attributted that redundant error steps often appear more "normal" or plausible than other types of erroneous steps, hindering the models' ability to identify them based on surface-level textual cues.
This suggests that current PRMs may be limited by their reliance on surface-level pattern recognition for error detection, highlighting the need for more profound reasoning and analytical capabilities.

\subsection{Detailed Analysis}

This section delves into a more nuanced analysis of our proposed \textsc{Socratic-PRMBench}, aiming to identify current models' limitations in providing process-level rewards and provide insights to guide the future development of PRMs.

\begin{figure}[t]
    \centering
    \includegraphics[width=0.9\linewidth]{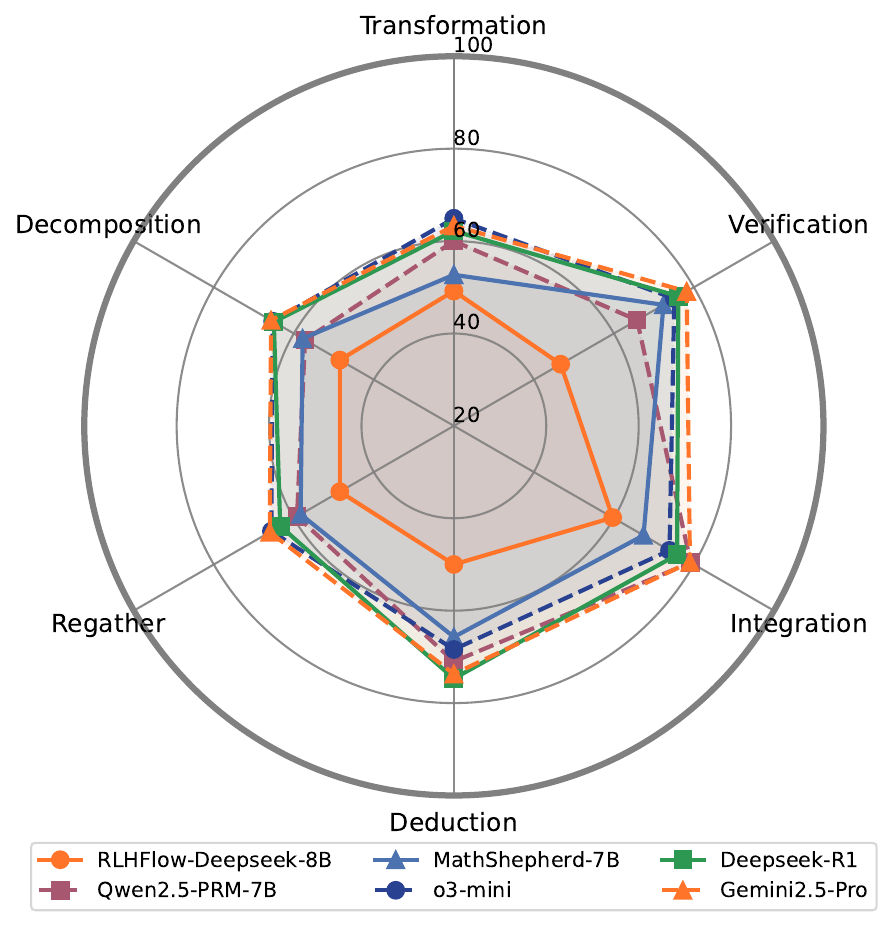}
    \caption{Average PRM-Score of representative PRMs and LLMs across 6 reasoning patterns. Both PRMs and LLMs shows imbalanced performance.}
    \vspace{-10pt}
    \label{fig:pattern_avg}
\end{figure}

\paragraph{Disparities in Performance across Reasoning Patterns} 
As shown in Figure \ref{fig:pattern_avg}, we present the average PRM-Scores of representative PRMs and LLMs across the six reasoning patterns.
A notable finding is the imbalanced performance exhibited by both PRMs and LLMs across different reasoning patterns. The performance of almost all models was consistently weaker on \textit{Transformation}, \textit{Decomposition}, and \textit{Regather} patterns compared to \textit{Deduction}, \textit{Integration}, and \textit{Verification}. 
This issue is more pronounced for PRMs, for example, Qwen2.5-Math-PRM-7B achieved a PRM-Score close to 80.0 on the \textit{Integration} pattern but struggles to reach 60.0 on the \textit{Decomposition} pattern.
This finding highlights a potential bias in the current PRM training data construction process. Existing PRM datasets, regardless of whether they're manually annotated or synthetically generated, appear to lack adequate representation of different reasoning patterns.
Due to the greater frequency of certain patterns like \textit{Deduction}, these datasets tend to be dominated by those patterns, resulting in significantly worse performance on rarer patterns such as \textit{Decomposition}. 
This observation underscores the importance of considering the distribution of different reasoning patterns in future PRM training data construction, as early detection of reasoning errors is critical to mitigate error propagation.

\begin{figure}[t]
    \centering
    \includegraphics[width=1.0\linewidth]{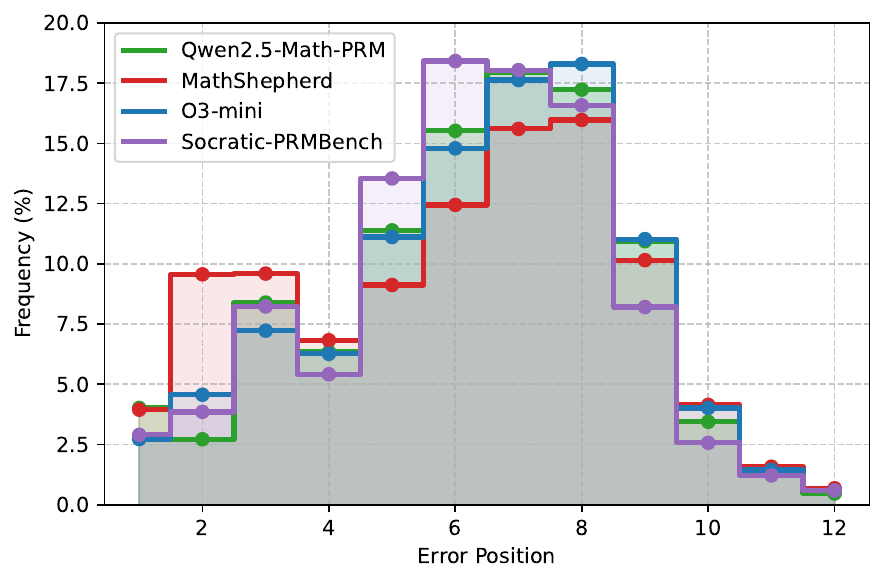}
    \caption{Error position distribution (truncated to 12) of   \textsc{Socratic-PRMBench} and the predicted error position distribution of several PRMs and LLMs.}
    \vspace{-10pt}
    \label{fig:err_pos_dist}
\end{figure}

\paragraph{Models Show Latency in Indentifying Error Steps}
To investigate the ability of models to detect reasoning errors in time, we compared the distribution of the ground truth error step positions in \textsc{Socratic-PRMBench} with the distributions of predicted error positions for representative PRMs and LLMs. 
As evidenced by Figure \ref{fig:err_pos_dist}, Qwen2.5-Math-PRM and o3-mini show a marked shift towards later steps compared to the ground truth distribution, indicating a delay in detecting early errors. 
This implies a limited ability to detect errors early on, allowing them to propagate. On the other hand, MathShepherd exhibits an opposite trend, with its predicted distribution shifts toward the beginning of the reasoning chain, suggesting that MathShepherd is prone to falsely identifying correct steps as errors, especially in the early stages of reasoning.
This inspires us that both early detection and avoidance of excessive false positives are crucial. 
Although propagation of errors will waste computational resources and reduces sampling efficiencys, overly aggressive error detection can prematurely terminate correct reasoning paths, hindering the exploration of potentially optimal solutions.

\begin{table}
\centering
\resizebox{0.95\linewidth}{!}{
    \begin{tabular}{lcccc}
    \toprule
    \multirow{2}{*}{\textbf{Model}}  & \multicolumn{3}{c}{\textbf{Accuracy}} & \multirow{2}{*}{\tabincell{c}{\textbf{PRM}\\ \textbf{Score} }}  \\
    \cline{2-4} 
    &Corr. & Err. & All. &\\
    \midrule
    \cellcolor{gray!10}Random $^\dagger$ &\cellcolor{gray!10}50.0 &\cellcolor{gray!10}50.0 &\cellcolor{gray!10}50.0&\cellcolor{gray!10}50.0 \\

    \hline \multicolumn{5}{c}{\textit{\textbf{Process Reward Models (PRMs)}}} \\   \hline 
    ReasonEval-7B & 87.3 & 35.7 & 69.6 & 61.9  \\
    Skywork-PRM-7B & 22.7 & 93.0 & 44.5 & 43.6 \\
    MathShepherd & 73.3 & 56.0 & 67.4 & 64.4 \\
    Qwen2.5-Math-PRM-7B & 90.8 & 42.9 & 74.5 & 68.0 \\
     \hline \multicolumn{5}{c}{\textit{\textbf{LLMs, Prompted as Critic Models}}} \\   \hline 
    GPT-4o & 83.0 & 57.5 & 74.6 & 70.8 \\
    QwQ-32B & 83.9 & 63.1 & 76.8 & 73.8  \\
    o3-mini & 82.6 & 69.0 & 78.0 & 75.7  \\
    Gemini-2.5-Pro & 83.6 & 62.8 & 76.5 & 73.5  \\
    
    \bottomrule
    \end{tabular}
}
\caption{Comparison of model performance on positive and negative test cases.$^\dagger$ represents performance of Random Guess.}
\label{tab:bias}
\vspace{-10pt}
\end{table}

\paragraph{Reward Bias of PRMs} Table \ref{tab:main_results} reveals that some PRMs perform even worse than random guessing, suggesting a substantial bias in their predictions. 
To further quantify this bias, we calculated accuracy for correct and error reasoning steps for each model. 
As shown in Table \ref{tab:bias}, the results reveal a clear reward bias within PRMs, with some models heavily favoring positive rewards and others tending to provide negative rewards. 
For instance, Qwen2.5-Math-PRM-7B displays a 90.8\% accuracy on correct steps but only a 42.9\% accuracy on error steps. 
In stark contrast, Skywork-PRM-7B shows a 93.0\% accuracy on error steps but only a 22.7\% accuracy on correct steps. 
While LLMs exhibits less pronounced bias than PRMs, however, a considerable gap remained in accuracy between correct and error steps.
Moreover, all the evaluated LLMs tended to favor positive rewards, which may limit their reliability in identifying subtle errors when serve as critic models.

\section{Conclusion}
In this work, we propose \textsc{Socratic-PRMBench}, a systematic and fine-grained benchmark for PRMs. \textsc{Socratic-PRMBench} comprises 2995 instances, categorized into six primary reasoning patterns and 20 sub-categories of fine-grained error types.
Through a systematic and comprehensive evaluations of existing PRMs and LLMs prompted as critic models, we observe potential shortcomings in existing models and provide valuable insights for future efforts on upgrading PRMs.

\section*{Limitations}
Although our work can provide a systematic and comprehensive evaluationg for PRMs,
the current version of our benchmark primarily focuses on reasoning tasks with objectively verifiable answers, such as mathematical problem. 
Applying our existing data construction methods to tasks in domains like literature, medicine, or law, where definitive ground truth is often absent, needs further exploration.
We intend to expand our benchmark to encompass a broader range of tasks in future versions of our benchmark.


\bibliography{anthology,custom}

\clearpage
\appendix

\section{Experimental Details}
\label{sec:appendix}
\paragraph{Abbreviation of Sub-Categories}
The full names of abbreviations used in our experiments are shown in Table \ref{apdxtab:abbrs}.

\begin{table}[h]
\centering
\resizebox{\linewidth}{!}{
    \begin{tabular}{lcc}
    \toprule
    \textbf{Abbr.} & \textbf{Full Name} & \textbf{Reasoning Pattern} \\
    
    \midrule
    TT. & Transformation Inconsistency & Tansformation \\
    TF. & Transformation Counter-Factuality & Transformation \\
    DC. & Decomposition Incompleteness & Decomposition \\
    DR. & Decomposition Redundancy & Decomposition \\
    DS. & Decomposition Unsoundness & Decomposition \\
    GP. & Regather Imprecision & Regather \\
    GC. & Regather Incompleteness & Regather \\
    GR. & Regather Redundancy & Regather \\
    CE. & Correction Error & Verification \\
    DE. & Detection Error & Verification \\
    CF. & Conclusion Counter-Factuality & Deduction \\ 
    CT. & Conclusion Inconsistency & Deduction \\ 
    CV. & Conclusion Invalidity & Deduction \\ 
    PC. & Premise Incompleteness & Deduction \\ 
    PR. & Premise Redundancy & Deduction \\ 
    PS. & Premise Unsoundness & Deduction \\ 
    IC. & Integration Incompleteness & Integration \\ 
    IT. & Integration Inconsistency & Integration \\ 
    IR. & Integration Redundancy & Integration \\ 
    IS. & Integration Unsoundness & Integration \\ 
    \bottomrule
    \end{tabular}
}
\caption{The full names of abbreviations.}
\label{apdxtab:abbrs}
\end{table}

\paragraph{Implementation Details}
For the training of Socratic reasoning model, we use LoRA tuning \citep{hu2021loralowrankadaptationlarge} to fine-tune a Qwen2.5-72B-Instruct with LLaMA-Factory library\footnote{https://github.com/hiyouga/LLaMA-Factory}.
For the evaluation of open-source PRMs, we utilize PRM Eval ToolKit\footnote{https://github.com/ssmisya/PRMBench} for implementation.
For the evalutation of LLMs prompted as critic models, we prompt LLMs with the prompt template in Table \ref{critic_prompt}, with default temperature set to 1.0.
During the test case construction procedure, we select $N=150$ samples from metadata set $D'$, including 10 samples from GSM8k and 50 samples each from Omni-Math, MathBench, and OlympiadBench.

\section{Prompts}
\label{appendix_prompt}
As described in Section \ref{sec:method}, LLMs play a crutial role in our method. 
In the socratic reasoning curation stage, the prompt for step-level verification is illustrated in Table \ref{prompt_step_verify}. 
In the test case construction stage, we follow \citep{song2025prmbench} and design task prompt and output format prompt seperately, as shown in Table \ref{task_prompt} and Table \ref{output_prompt} respectively.
For the LLM-based filering procedure, we use the prompt template in Table \ref{filter_prompt}.

\begin{table*}[h]
\begin{tabularx}{\linewidth}{X}
\toprule 
\textbf{Prompt Template for Evaluation of LLMs prompted as critic models} \\
\cmidrule{1-1}
\textbf{[System Prompt]}\\
You are a mathematical reasoning evaluator. Your task is to analyze mathematical problem-solving steps and provide structured assessments in JSON format.\\
\\
For each solution step, you need to evaluate its Validity Score (-1 to +1):\\
   * +1: Completely correct mathematical reasoning\\
   * 0: Partially correct with some mistakes\\
   * -1: Completely incorrect\\
   * Use any value in between to indicate varying degrees of correctness\\
\\
Requirements:\\
- Evaluate each step independently\\
- Provide scores as floating-point numbers\\
- Return results in strict JSON format: \{\textquotedbl validity \textquotedbl: [scores]\} \\
- Ensure the array have the same length as the number of steps\\
- Maintain mathematical rigor in your evaluation\\
- Consider mathematical accuracy, logical coherence, and solution efficiency\\
\\
Example output format:\\
\\
\{\textquotedbl validity \textquotedbl: [0.8, -0.5, 1.0]\}\\
\\
You will be presented with a mathematical problem and its step-by-step solution. Please analyze each step and provide your evaluation in the specified JSON format.
\\
\\
\textbf{[User]}\\
Question: \{question\} \\
\\
Solutions: \{solution\} \\
\bottomrule

\end{tabularx}
\caption{Prompt template for evaluation of LLMs prompted as critic models}
\label{critic_prompt}
\end{table*}

\begin{table*}[h]
\begin{tabularx}{\linewidth}{X}
\toprule 
\textbf{Prompt Template for Step Verification} \\
\cmidrule{1-1}
You are an expert on reasoning process verification, you will be given a question, a solution(split into paragraphs, enclosed with tags and indexed from 1, and a reference answer.\\
\\{}
\textbf{[Question]} \\
\{question\} \\
\\{}
\textbf{[Solution]} \\
\{solution\} \\
\\{}
\textbf{[Reference Answer]} \\
\{answer\}\\
\\
Your task is to review and critique the solution paragraph by paragraph. Once you identify an error in a paragraph, return the index of the paragraph where the earliest error occurs. Otherwise, return the index of -1 (which typically denotes "not found"). Please put your final answer (i.e., the index) in \textbackslash boxed\{\}.
\\
\bottomrule

\end{tabularx}
\caption{Prompt template for step verification.}
\label{prompt_step_verify}
\end{table*}

\begin{table*}[h]
\begin{tabularx}{\linewidth}{X}
\toprule 
\textbf{Task Prompt for Test Case Construction} \\
\cmidrule{1-1}
You are a helpful AI assistant that is very good at reasoning and data construction. Now I want to test the ability of process-level reward models to judge whether a step within reasoning process is correct. To do this, please help me build flawed cases by introducing specific types of errors into a given reasoning process.\\
You will be provided with:\\
1. A mathematical problem.\\
2. A correct step-by-step reasoning process used to solve it. Each step is in a form of Action, posssibly including [Transformation], [Decomposition], [Regather], [Deduction], [Verification], [Integration], [Answer], [LVerification] and [GVerification].\\
\\
The description of Actions are as follows:\\
\\
\#\# [Transformation] (Identifier: <Repeat>xxx</Repeat>) \\
- Explain the problem from a problem-solving perspective \\
- Gain a more comprehensive and clear understanding of the problem through rephrasing \\
\\
\#\# [Decomposition] (Identifier: <Decomposition>xxx</Decomposition>)\\
- Break down the problem into several core sub-problems; resolve the main problem by tackling each sub-problem\\
- If no breakdown is necessary, provide the solution approach\\
\\
\#\# [Regather] (Identifier: <Regather>xxx</Regather>)\\
- Collect key information from the input relevant to solving the problem\\
- Output definitions, principles, and other concepts related to solving the problem, and provide explanations\\
\\
\#\# [Deduction] (Identifier: <Deduction>xxx</Deduction>)\\
- Observe existing information and extract key parts\\
- Identify explicit and implicit requirements, considering constraints and limitations \\
- Propose concrete ideas for solving the problem \\
- Execute reasoning according to the ideas \\
\\
\#\# [LVerification]\&[GVerification] (Identifier: <L(G)Verification>xxx</L(G)Verification>) \\
- Verify the logical consistency of the reasoning process \\
- Check the reasoning process against existing evidence \\
- Look for potential flaws in the reasoning process and refine them \\
- Review the completeness of understanding \\
- Question your assumptions and consider alternative viewpoints \\
- [LVerification] may occur after any reasoning step, verifying local steps \\
- [GVerification] only occurs between [Integration] step and [Answer] step, verifying global process \\
\\
\#\# [Integration] (Identifier: <Integration>xxx</Integration>) \\
- Integrate all current reasoning processes to form the current conclusion \\
\\
\#\# [Answer] (Identifier: <Answer>xxx</Answer>)\\
- Output the final answer to the original problem\\
\\
Your task is to modify the question, adjust original steps, or introduce additional steps into the original process chain to create a reasoning process that appears plausible but is incorrect, which leads to a wrong answer. The objective is to simulate flawed solutions by incorporating the specified error detailed after '\#\#\# Error Type to Introduce'.\\
\\
\#\#\# Error Type to Introduce\\
\{Error type\}
\\
\bottomrule

\end{tabularx}
\caption{Task prompt for test case construction.}
\label{task_prompt}
\end{table*}

\begin{table*}[h]
\begin{tabularx}{\linewidth}{X}
\toprule 
\textbf{Output Format Prompt for Test Case Construction} \\
\cmidrule{1-1}
\#\#\# Formatting Instructions:\\
\\
After making the modifications, provide the following structured output:\\
\{ \\
    \quad \textquotedbl original\_question\textquotedbl: \textquotedbl The original mathematical problem.\textquotedbl, \\
    \quad \textquotedbl modified\_question \textquotedbl: \textquotedbl The modified problem or original problem, \textquotedbl \\
    \quad  \textquotedbl original\_process\textquotedbl: [\textquotedbl original\_step 1 \textquotedbl, \textquotedbl original\_step 2\textquotedbl, \ldots], \\
    \quad \textquotedbl modified\_process\textquotedbl: [\textquotedbl modified\_step 1\textquotedbl, \textquotedbl modified\_step 2 \textquotedbl, \ldots], \\
    \quad \textquotedbl modified\_steps\textquotedbl: [1, 5, 7, \ldots], \\
    \quad \textquotedbl error\_steps\textquotedbl: [5, 6, \ldots], \\
    \quad \textquotedbl reason\textquotedbl: \textquotedbl Explanation for the changes.\textquotedbl \\
    
\} \\
\\
Detailed Requirements:\\
1. original\_question: A string representing the original mathematical problem as provided.\\
2. modified\_question: A string representing the modified problem after your changes. If the problem remains the same, you can copy the original question.\\
3. original\_process: A non-empty list of strings representing the original reasoning steps provided as input.\\
4. modified\_process: A non-empty list of strings representing the reasoning process after your modifications. \\
5. modified\_steps: A non-empty list of integers indicating the indexes of all modified steps. Indexing starts at 1.\\
6. error\_steps: A non-empty list of integers representing the steps that contain hallucinations or errors. These should also be part of modified\_steps.\\
7. reason: A clear explanation of the modifications made, why they were introduced, and how they align with the specified error types. \\
\\
\#\#\# Notes:\\
1. Ensure all lists are non-empty.\\
2. Use LaTeX format for all mathematical symbols (e.g., $x^2$ for $x$ squared). Do not use Unicode symbols such as \textbackslash u2248 or \textbackslash u00f7.\\
3. Ensure the JSON object is well-formed, with proper escaping for special characters like backslash n (e.g., use backslash backslash n for newlines).\\
4. All indexes start from 1, that is, the first step's index is 1, not 0.\\
5. You can choose to modify the question or not, if the question remains the same, you can copy the original question. But if the question is modified, ensure that the steps is judged based on the \\modified question.\\
6. Please give original process as provided by the prompt, do not modify it. \\
\bottomrule

\end{tabularx}
\caption{Output format prompt for test case construction.}
\label{output_prompt}
\end{table*}

\begin{table*}[h]
\begin{tabularx}{\linewidth}{X}
\toprule 
\textbf{Prompt Template for LLM-based Filtering} \\
\cmidrule{1-1}
You are an expert on reasoning process verification, you will be given a question, a solution (split into paragraphs, enclosed with tags.\\
\\
Your task is to decide whether the step-by-step solution generated by LLMs satisfies:\\
1. The process generated by LLMs seems like a possible solution path that could happen.\\
2. The process generated by LLMs is exactly wrong and the type of error is suitable for the description of [classification]\\
\\{}
\textbf{[Classification]} \\
\{classification\}\\
\\{}
\textbf{[Question]} \\
\{question\}\\
\\{}
\textbf{[Solution]} \\
\{Solution\} \\
\\
Please answer a “Yes” if both of the two aspects are satisfied, otherwise answer 'No'.\\
Please put your final answer (Yes or No) in \textbackslash boxed \{\}.
\\
\bottomrule

\end{tabularx}
\caption{Prompt template for LLM-based Filtering}
\label{filter_prompt}
\end{table*}


\end{document}